\documentclass[twoside,twocolumn,10pt]{article}

\usepackage{wscg}           
\usepackage{soul}
\usepackage{makecell}

\RequirePackage{ifpdf}
\ifpdf
 \RequirePackage[pdftex]{graphicx}
 \RequirePackage[pdftex]{color}
\else
 \RequirePackage[dvips,draft]{graphicx}
 \RequirePackage[dvips]{color}
\fi
\usepackage{xcolor}
\usepackage[english]{babel}     
\usepackage{array}          
\usepackage[intlimits]{amsmath} 
\usepackage{hyphenat}      
\usepackage{amsfonts}
\usepackage{multirow}

\usepackage{stackengine}

\usepackage{comment}
\usepackage{subcaption}
\usepackage[breaklinks]{hyperref}
\usepackage{doi}
\usepackage{csquotes}
\usepackage{adjustbox}
\usepackage{graphicx} 

\usepackage[style=alphabetic]{biblatex}
\DeclareLabelalphaTemplate{
  \labelelement{
    \field[final]{shorthand}
    \field{label}
    \field[strwidth=3,strside=left,names=1]{labelname}
  }
  \labelelement{
    \field[strwidth=2,strside=right]{year}    
  }
}

\graphicspath{{figures/}} 
\usepackage{nopageno}       
\usepackage[switch,columnwise]{lineno}

\title{Examination of PCA Utilisation for Multilabel Classifier of Multispectral Images}

\author{
Filip Karpowicz, Wiktor Kępiński, Bartosz Staszyński, Grzegorz Sarwas \\
Warsaw University of Technology, Faculty of Electrical Engineering \\
Pl. Politechniki 1, 00-661 Warsaw, Poland \\
\{filip.karpowicz.stud, wiktor.kepinski.stud, bartosz.staszynski.stud, grzegorz.sarwas\}@pw.edu.pl
}

\usepackage{url}
\makeatletter
\def\Uslash{\mathbin{\mathchar`\/}\@ifnextchar{/}{\kern-.15em}{}}
\g@addto@macro\UrlSpecials{\do \/ {\Uslash}}
\def\Ucolon{\mathbin{\mathchar`:}\@ifnextchar{/}{\kern-.1em}{}}
\g@addto@macro\UrlSpecials{\do : {\Ucolon}}
\makeatother

\begin{document}

\twocolumn[{\csname @twocolumnfalse\endcsname

\maketitle  

\begin{abstract}
\noindent
This paper investigates the utility of Principal Component Analysis (PCA) for multi-label classification of multispectral images using ResNet50 and DINOv2, acknowledging the high dimensionality of such data and the associated processing challenges. Multi-label classification, where each image may belong to multiple classes, adds further complexity to feature extraction. Our pipeline includes an optional PCA step that reduces the data to three dimensions before feeding it into a three-layer classifier. The findings demonstrate that the effectiveness of PCA for multi-label multispectral image classification depends strongly on the chosen deep learning architecture and training strategy, opening avenues for future research into self-supervised pre-training and alternative dimensionality reduction approaches.

\vspace{0.5em}

\subparagraph{Keywords:}
PCA, Feature Extraction, Multilabel Classifier, Multispectral Images, Soft Contrastive Learning

\vspace*{1.0\baselineskip}

\end{abstract}
}]

\section{Introduction}
Multispectral imaging extends the capabilities of traditional RGB photography by capturing data across multiple discrete electromagnetic spectrum bands. Such images, which often include visible, near-infrared, and occasionally thermal bands, can include four to twenty spectral channels. This expanded spectral range enables the detection of features imperceptible to the human eye, thereby providing invaluable insights in fields such as precision agriculture, environmental monitoring, and urban planning, where subtle spectral variations may reveal information about crop stress, pollution levels, or material composition~\cite{RAM2024109037,rs12152503}. 

In contrast to conventional single-label classification, multi-label classification captures the inherent complexity of multispectral imagery by allowing each pixel or region to belong to multiple classes simultaneously. For example, a single pixel may exhibit characteristics of both ‘forest’ and ‘flooded vegetation,’ thus enabling a more nuanced representation of real-world scenarios~\cite{softcon}.

\begin{figure}[b!]
    \noindent\fbox{%
        \parbox{\dimexpr\linewidth-2\fboxsep-2\fboxrule\relax}{%
           \small{Permission to make digital or hard copies of all or part of this work for personal or classroom use is granted without fee provided that copies are not made or distributed for profit or commercial advantage and that copies bear this notice and the full citation on the first page. To copy otherwise, or republish, to post on servers or to redistribute to lists, requires prior specific permission and/or a fee.}
        }%
    }
\end{figure}

This paper focuses on the SSL4EO (Self-Supervised Learning for Earth Observation) dataset~\cite{ssl4eodata}, which comprises imagery from multiple satellite platforms, including Sentinel-1, Sentinel-2, and Landsat. These platforms provide complementary information through different sensing modalities, such as optical, near-infrared, and synthetic aperture radar (SAR).

Multispectral images obtained from satellites often exhibit high correlation among spectral channels, introducing a degree of redundancy in data processing. The dimensionality of the data poses a significant challenge, contributing to increased processing costs. One potential solution is to reduce the dimensionality, thereby minimising redundancy and lowering the overall processing cost. Principal Component Analysis (PCA) is widely used for this purpose, achieving optimal results by capturing most of the data variance in the initial principal components, while ensuring complete data decorrelation by maximising the variance of the output data. 

A common approach involves leveraging pre-trained deep learning architectures such as ResNet (Residual Networks) and Vision Transformers (ViT) for feature extraction from multispectral imagery. These models, originally developed for natural image classification, have been adapted to remote sensing tasks via contrastive learning, followed by transfer learning and fine-tuning. ResNet’s skip connections enable efficient training of very deep networks, allowing the extraction of hierarchical features at multiple scales. Meanwhile, ViT’s attention mechanisms capture long-range dependencies in the data, making it particularly effective for understanding spatial relationships in multispectral imagery. Such architectures are widely adopted because of their proven ability to learn robust, generalisable features that outperform handcrafted methods, especially when handling the complex spectral and spatial correlations present in multispectral satellite data~\cite{7514991,kim2023visiontransformerbasedfeatureextraction,LIAO2025105036,dimensionredu}.

This study investigates using Principal Component Analysis (PCA) for dimensionality reduction on multispectral images, testing its utility within a multi-label classification pipeline employing ResNet50 or DINOv2 feature extractors. We conclude that PCA's utility is highly architecture-dependent, benefiting fine-tunable transformers but not CNNs or frozen models in this context, while also confirming PCA's advantage in reducing model size and potentially inference time.

\section{Related Work}
In recent years, the rapid growth of remote sensing data has driven significant advances in machine learning techniques for Earth observation. Self-supervised learning, contrastive learning, and dimensionality reduction methods have become key approaches for extracting meaningful representations from large volumes of unlabeled and high-dimensional satellite imagery. Numerous datasets and algorithms have been proposed to improve the efficiency and accuracy of geospatial data analysis, enabling applications such as land cover classification, environmental monitoring, and urban planning. This section reviews relevant works that have laid the foundation for current research in these areas.

\subsection{SSL4EO-S12}
The SSL4EO-S12 (Self-Supervised Learning for Earth Observation—Sentinel-1/2) dataset is a large-scale, globally distributed, multimodal, and multitemporal dataset designed to advance self-supervised learning in Earth observation~\cite{ssl4eodata}. It addresses limitations found in earlier datasets, such as restricted geographical diversity, limited temporal coverage, and a lack of multimodal integration.

SSL4EO-S12 comprises three million image patches, each covering an area of $2640\times2640$ metres, sampled from 250,000 unique locations worldwide. Each location includes four seasonal snapshots, ensuring that the dataset captures a wide variety of climatic and land-cover conditions. The dataset integrates Sentinel-1 GRD (Synthetic Aperture Radar imagery with VH and VV polarisations), Sentinel-2 L1C (top-of-atmosphere multispectral images), and Sentinel-2 L2A (surface reflectance multispectral images with atmospheric correction), providing a rich multimodal representation for remote sensing applications. A geospatial sampling strategy ensures global coverage, with locations drawn from the 10,000 most populated cities. This approach balances urban and rural environments while minimising redundancy through overlap filtering techniques.

Publicly available satellite data from the European Space Agency’s Sentinel-1 and Sentinel-2 missions were used to compile the dataset. A structured methodology was employed to ensure high-quality and diverse samples. Locations were selected using a Gaussian distribution centred on major cities, thus providing a representative mix of land-cover types. To maintain high optical image quality, Sentinel-2 images were filtered to include only those with less than 10\% cloud coverage. Additionally, all Sentinel-1 and Sentinel-2 images were aligned to a uniform spatial resolution to facilitate multimodal learning. To further enhance the dataset’s utility, overlap filtering was applied to ensure that selected image patches did not significantly overlap, thereby avoiding redundant samples. The dataset structure also includes extensive metadata—such as geographic coordinates, acquisition time, sensor type, and processing level—allowing for adaptation to a wide range of domain-specific tasks.

Experiments show that models pre-trained on SSL4EO-S12 outperform those trained solely on labelled datasets, demonstrating superior feature extraction capabilities. Nevertheless, recent work in contrastive learning (e.g. SupCon, SoftCon) indicates that introducing labelled data into contrastive learning can further boost model performance. In this context, Yi Wang et al.~\cite{softcon} have built a labelled Earth observation dataset named SSL4EO-S12-ML, aggregating the original SSL4EO-S12 dataset with land-cover data from Google Dynamic World. The data are classified into nine categories: water, trees, grass, flooded vegetation, crops, shrub and scrub, built, bare, and snow and ice, as shown in Table~\ref{tab:land-cover-classes}.

\begin{table}[htbp]
  \centering
  \begin{tabular}{|c|l|}
    \hline
    \textbf{Class ID} & \textbf{Land Cover Type} \\
    \hline
    0 & Water \\
    1 & Trees \\
    2 & Grass \\
    3 & Flooded vegetation \\
    4 & Crops \\
    5 & Shrub and scrub \\
    6 & Built \\
    7 & Bare \\
    8 & Snow and ice \\
    \hline
  \end{tabular}
  \caption{Land Cover Classification}
  \label{tab:land-cover-classes}
\end{table}

\subsection{Softcon}
Soft Contrastive Learning (SoftCon) enhances traditional contrastive learning by introducing a soft, similarity-based approach to more effectively handle multi-label satellite imagery~\cite{softcon, 9875399}. The key idea behind SoftCon is to ensure smooth transitions in the feature space by pulling together image representations according to their degree of semantic similarity rather than relying on strictly binary positive or negative assignments. In traditional self-supervised contrastive learning methods, all images (apart from an augmented version of the anchor image) are treated as negatives~\cite{khosla2021supervisedcontrastivelearning}. By contrast, SoftCon computes similarity between images based on their multi-label annotations, specifically using cosine similarity between multi-hot label vectors, thus providing a more nuanced definition of positive pairs.

SoftCon introduces a novel loss function that aligns feature similarity with label similarity. Instead of treating samples as strictly positive or negative, the model is encouraged to learn a latent space in which the similarity of feature representations reflects the similarity of their respective label distributions. The loss function is defined as:
\begin{equation}
L = L_{Contrast} + \lambda * L_{SoftCon},
\end{equation}
where
\begin{equation}
L_{Contrast} = - \sum_{i=1}^{N} \log \frac{\exp \left( z_i \cdot z_i' / \tau \right)}{\sum_{j=1}^{N} \exp \left( z_i \cdot z_j' / \tau \right)}
\end{equation}
and
\begin{align}
L_{SoftCon} = &-\sum_{i=1}^{N}\sum_{j=1}^{N} \left( Y_{ij} \log\sigma\left(X_{ij}\right) \right. \nonumber  \\ 
  & +  \left(1 - Y_{ij}\right) \log\left(1 - \sigma\left(X_{ij}\right)\right), 
\end{align}
where  $Y_{ij} = y_i \otimes y_j^T$, $X_{ij} = z_i \otimes z_j^T$, $z_i$ and $z_i'$ are the feature vectors of the $i$-th image and its augmentation, respectively. $y_i$ is the multi-hot vector corresponding to the $i$-th image. $\lambda$ is a weighing parameter for the soft contrastive loss, $\tau$ is the softmax temperature and $\sigma(\cdot)$ is the sigmoid function. Feature vectors are a numerical representation of a set of characteristics or features of an object, in our case, satellite images. The feature vector is the output of an encoder (a neural network trained to extract features from input data).

To further improve training efficiency for ViT backbones, SoftCon employs a Siamese masking strategy inspired by Masked Autoencoders (MAE)~\cite{he2021masked,chen2020exploringsimplesiameserepresentation}. During training, some input patches are randomly masked in one of the network branches, thereby reducing computational complexity while preserving essential semantic information.

Studies have shown that SoftCon-based models outperform many models trained with traditional self-supervised learning across multiple downstream tasks. This establishes SoftCon as the current state-of-the-art Earth observation foundation model, making it a valuable reference for our approach.

\subsection{PCA}
Principal Component Analysis (PCA) is a widely used technique for reducing the dimensionality of high-dimensional datasets while preserving as much variance as possible~\cite{9206927}. The process begins with data standardisation, ensuring that all features contribute equally to the analysis. This step involves subtracting the mean and dividing by the standard deviation for each feature, transforming the dataset into a standardised form where all variables have a mean of zero and a standard deviation of one.

Standardisation is crucial because PCA is sensitive to the scale of variables, and unstandardised data can lead to misleading results. Next, the covariance matrix is computed to examine relationships between variables. The covariance matrix captures the degree to which different features vary, providing insight into how strongly variables correlate. A high absolute covariance value between two features indicates a strong linear relationship, whereas a near-zero value suggests little to no correlation. PCA aims to transform the original correlated features into a new set of uncorrelated features, called principal components, which are ordered by the amount of variance they explain in the data.

To achieve this transformation, eigenvalue decomposition is applied to the covariance matrix. This step identifies the eigenvectors and their corresponding eigenvalues, where the eigenvectors represent the principal components, and the eigenvalues indicate the amount of variance captured by each component. The eigenvectors define a new orthogonal basis in which the first component captures the maximum variance in the data, the second captures the highest variance while remaining orthogonal to the first, and so on. The eigenvalues indicate the amount of variance preserved along each eigenvector direction.

The number of retained principal components is determined based on cumulative explained variance, typically by selecting the top components that account for a sufficient proportion (e.g., 95\%) of the total variance.

Finally, dimensionality reduction is performed by projecting the original data onto the selected principal components. This transformation results in a new dataset with fewer dimensions, retaining the most essential information from the original data. The reduced representation enables more efficient data processing and visualisation while mitigating challenges associated with high-dimensional data, such as overfitting and computational complexity. PCA is extensively applied in areas such as image compression, feature extraction, and noise reduction, making it an essential tool in data science and machine learning~\cite{shlens2014tutorialprincipalcomponentanalysis}.

\subsection{PCA in Multispectral Imagery}


The application of Principal Component Analysis (PCA) in multispectral imagery offers several benefits that enhance the efficiency and quality of data analysis. One of its primary advantages is reducing dimensionality, transforming a large set of correlated spectral bands into a smaller, more manageable number of uncorrelated principal components. This reduction significantly lowers computational demands, making it feasible to process extensive datasets without sacrificing essential information, as the retained components typically account for over 95\% of the data’s variance. Furthermore, PCA acts as an effective noise filter by discarding lower-variance components, which often capture random fluctuations rather than meaningful patterns, thereby improving the signal-to-noise ratio and focusing attention on critical features such as land cover types or vegetation health. This denoising capability can lead to enhanced accuracy in subsequent classification tasks. PCA also facilitates data visualisation by compressing multidimensional information into a few principal component images, enabling researchers to identify dominant patterns—such as overall brightness in the first component or specific spectral contrasts in later ones—without the clutter of redundant bands. These benefits make PCA a powerful preprocessing tool, streamlining analysis while preserving the essence of multispectral imagery for applications ranging from environmental monitoring to urban planning~\cite{Zheng-2021}.

\section{Proposed Solution}
In our study, we employ ResNet and DINOv2 as feature extractors in two separate experiments to evaluate how different architectures process multispectral images. Both models act as encoders within our PCA → Encoder → Classifier pipeline, yet they diverge significantly in their approaches to feature extraction and representation learning.

ResNet, a convolutional neural network (CNN), extracts features through a hierarchical structure of convolutional layers. Each layer applies localised filters to detect patterns such as edges, textures, and more complex shapes. The network utilises residual connections, which help preserve information flow across layers, enabling deeper feature extraction without gradient degradation~\cite{he2015deepresiduallearningimage}. ResNet processes the input image by progressively refining feature maps, where early layers capture fine-grained spatial details, and deeper layers extract high-level semantic features. Finally, a global average pooling (GAP) layer aggregates spatial information into a single feature vector, providing a compressed representation of the image and highlighting its most relevant characteristics for classification.

By contrast, DINOv2, a vision transformer (ViT)-based model, employs a fundamentally different approach to feature extraction~\cite{oquab2024dinov2learningrobustvisual}. Rather than applying convolutional filters, it divides the image into non-overlapping patches, which are then embedded into high-dimensional vectors. These patch embeddings are processed through multiple self-attention layers, allowing the model to analyse relationships between different parts of the image, even those that are spatially distant. Unlike ResNet, which primarily captures local features, DINOv2 constructs a global image representation, making it particularly effective at discerning long-range dependencies and spectral variations across bands. The model outputs a set of feature tokens, each corresponding to different aspects of the image, rather than a single condensed feature vector. This results in a richer, more flexible representation, with each token contributing distinct information for the classification task.

The key difference between these two architectures lies in how they extract and represent features. ResNet processes an image through a sequence of convolutional layers, gradually refining spatial features before condensing them into a single vector, which serves as a compact representation of the image. In contrast, DINOv2 applies self-attention mechanisms across patch-based embeddings, producing multiple feature tokens that capture a more context-aware and globally informed representation.

In separate experiments, we use these two models as encoders to analyse how their feature representations interact with dimensionality reduction via PCA. This approach will enable us to determine whether PCA enhances feature extraction for spatially focused CNN-based representations, globally distributed transformer-based representations, or both, ultimately shedding light on the most effective method for multispectral image classification.

We adopt state-of-the-art baseline models with encoder weights pretrained on SoftCon and ImageNet datasets~\cite{5206848}. These models provide robust benchmarks for assessing the impact of PCA in our study.

\subsection{PCA Model}
In our analysis, a Principal Component Analysis (PCA) model was implemented using the Scikit-learn library in Python~\cite{scikit-learn}. Due to computational constraints, the model was trained on 40\% of randomly selected samples from the dataset, accounting for approximately 100,430 samples. Each sample is composed of 13 bands. This sampling percentage represented the maximum volume our hardware could effectively process. The resulting PCA model demonstrated excellent data representation capabilities, with the components collectively explaining 99.9\% of the total variance in the data, indicating that the PCA transformation successfully captured the essential features of the image dataset while substantially reducing dimensionality.

\subsection{Classifiers}
In our study, we employ a three-layer fully connected classifier with a simple feedforward architecture to perform multi-label classification on the extracted features. This classifier receives feature embeddings from either ResNet or DINOv2 and processes them via a series of linear transformations and non-linear activations to generate classification outputs. We deliberately keep the design of this classifier minimal to avoid introducing unnecessary complexity, thus ensuring that our evaluation focuses primarily on the effects of PCA and the feature extraction process. This straightforward approach helps prevent any undue influence on classification performance.

We include dropout layers between the hidden layers to mitigate overfitting, starting with a dropout rate of 0.4 and gradually decreasing with each deeper layer. Finally, the output layer maps the processed features to the target classes.


\subsection{Pipeline}
\begin{figure}[h]
    \centering
    \includegraphics[width=\linewidth]{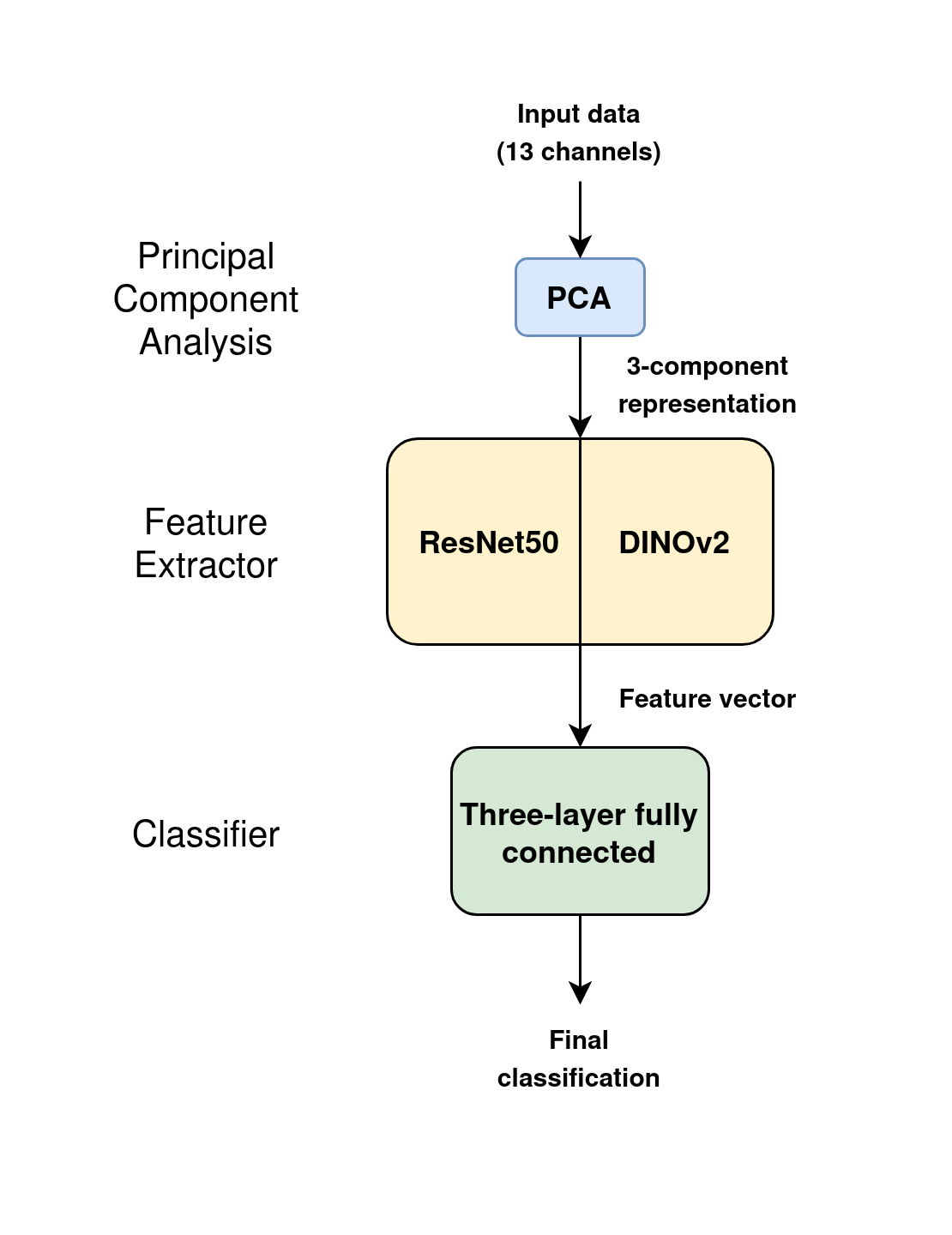}
    \caption{Suggested pipeline}
    \label{fig:pipeline}
\end{figure}

Our experimental pipeline (Figure~\ref{fig:pipeline}) comprises three main stages: Principal Component Analysis (PCA), a feature extractor (ResNet50 or DINOv2), and a three-layer classifier. First, PCA is applied to the input data to reduce dimensionality while preserving essential variance, resulting in a three-dimensional representation that serves as input for the feature extractor. To ensure compatibility with this transformed input, the first layer of each encoder is replaced and randomly initialised, enabling it to process a three-dimensional input, rather than the original 13-dimensional input, which the SoftCon weights are adapted to. The feature extraction stage is then performed using either ResNet50 or DINOv2, and the extracted features are passed to a fully connected, three-layer classifier for final predictions. Since ResNet50 and DINOv2 produce different feature outputs, the classifier’s input dimensions vary accordingly, leading to differences in the number of neurons in the hidden layers. Finally, a sigmoid function is applied to the classifier’s output vector, producing the final classification results.

\section{Experiments}
We conducted experiments to evaluate the effectiveness of PCA under various conditions and across different architectures. The following metrics served as indicators:
\begin{itemize}
    \item  model size,
    \item  model inference time,
    \item  model accuracy,
    \item  model recall,
    \item  model precision,
    \item  model F1-score.
\end{itemize}

To better isolate the effect of PCA on classification performance, we carried out a series of experiments with a range of model configurations, as shown in Table~\ref{tab:experiment_conf}. Each configuration was evaluated both with and without PCA, allowing for a direct comparison of its impact on the model’s performance.
\begin{table}[!ht]
    \centering
    
    \begin{adjustbox}{width=\linewidth}
    \begin{tabular}{|c|c|c|c|}
        \hline
         Preprocessor & Encoder & Weights Source & \makecell{Encoder \\ fine-tuning}  \\
         \hline
         \multirow{8}{*}{PCA/None} & \multirow{4}{*}{resnet50} & \multirow{2}{*}{softcon} &  finetuned \\
         \cline{4-4}
         & & & frozen* \\
         \cline{3-4}
         & & \multirow{2}{*}{imagenet} & finetuned \\
         \cline{4-4}
         & & & frozen \\
         \cline{2-4}
         & \multirow{4}{*}{dinov2} & \multirow{2}{*}{softcon} &  finetuned \\
         \cline{4-4}
         & & & frozen* \\
         \cline{3-4}
         & & \multirow{2}{*}{imagenet} & finetuned \\
         \cline{4-4}
         & & & frozen \\
         \hline
    \end{tabular}
    \end{adjustbox}
    \caption{Experimental Configurations \\ \footnotesize * Encoder weights were not updated during training except first layer}
    \label{tab:experiment_conf}
\end{table}

\subsection{Methodology}
The SSL4EO-S12-ML dataset was divided into training, validation, and testing sets in a ratio of 5:1:1.

Models were built according to the experimental configurations listed in Table~\ref{tab:experiment_conf}. It is worth noting that ImageNet encoder weights were adapted to a three-channel input, whereas SoftCon encoder weights were adapted to a thirteen-channel input. Because of this, there are configurations in which the source weights of the encoder cannot be set for the input layer. These cases are marked with an asterisk, indicating that the first layer of the encoder is left trainable.

The PCA preprocessing layer was not fine-tuned during model training. The PCA model was fitted separately by analysis of the training images alone.

Each model was combined with a three-layer classifier and evaluated using a Sigmoid-based Binary Cross-Entropy loss function. The Adam optimiser was used in conjunction with a learning rate scheduler, which lowers the learning rate upon plateau detection with a patience of two epochs and a reduction factor of 0.1. The models were deployed and trained on two GPUs using CUDA. Each model was trained for 100 epochs, with an early stopping mechanism with patience set to five epochs. The training dataset was sampled in batches of 64 images.

To create a benchmark for these experiments, each model was tested for average inference time, accuracy, recall, and precision.

To determine inference times, 2,510 samples were passed through each model five times, and the average inference time per sample was calculated.

\subsection{Results}
Table \ref{tab:inference} summarizes inference times per batch (64 samples) for ResNet50 and DINOv2 encoders, comparing performance with and without PCA layer across minimum, average, and maximum recorded times.

\begin{table}[!ht]
    \centering    
    \begin{adjustbox}{width=\linewidth}
    \begin{tabular}{|c|c|c|c|}
        \hline
         Encoder type & Results & With PCA & Without PCA  \\
         \hline
         \multirow{3}{*}{ResNet50} & Min & \textbf{45.28ms} & 45.45ms \\
         & Average & \textbf{45.80ms} & 47.47ms \\
         & Max & \textbf{46.83ms} & 52.69ms\\
         \hline
         \multirow{3}{*}{DINOv2} & Min & 96.05ms & 94.81ms \\
         & Average & 98.57ms & 97.03ms \\
         & Max & 103.99ms & 100.62ms \\
         \hline
    \end{tabular}    
    \end{adjustbox}
    \caption{Inference times per batch of 64 samples}
    \label{tab:inference}
\end{table}

The number of parameters and the model size depend on the chosen architecture. The results are presented in Table~\ref{tab:size_comprehension}.

\begin{table}[!ht]
    \centering
    \begin{adjustbox}{width=\linewidth}
    \begin{tabular}{|c|c|c|c|}
        \hline
         \makecell{Encoder\\type} & Results & With PCA & Without PCA  \\
         \hline
         \multirow{2}{*}{ResNet50} & Parameters & $\mathbf{26.135 \cdot 10^6}$ & $26.167 \cdot 10^6$ \\
         & Model size & \textbf{99.90MB} & 100.02MB \\
         \hline
         \multirow{2}{*}{DINOv2} & Parameters & $\mathbf{86.951  \cdot 10^6}$ & $88.456  \cdot 10^6$ \\
         & Model size & \textbf{331.69MB} & 337.43MB \\
         \hline
    \end{tabular}    
    \end{adjustbox}
    \caption{Size Comprehension}
    \label{tab:size_comprehension}
\end{table}
\begin{table}[!ht]
    \centering
    \begin{adjustbox}{width=\linewidth}
    \begin{tabular}{|c|c|c|c|}
        \hline
         Experiment & Results & With PCA & Without PCA  \\
         \hline
         \multirow{4}{*}{\makecell{ResNet50/\\SoftCon/\\fine-tuned}} & Accuracy & 83.82\% & 86.20\% \\
         & Precision & 76.32\% & 80.20\% \\
         & Recall & 71.92\% & 76.87\% \\
         & F1-Score & 71.41\% & 77.37\% \\
         \hline
         \multirow{4}{*}{\makecell{ResNet50/\\SoftCon/\\frozen}} & Accuracy & 83.91\% & 85.64\% \\
         & Precision & 77.89\% & 78.15\% \\
         & Recall & 70.88\% & 76.93\% \\
         & F1-Score & 71.94\% & 76.71\% \\
         \hline
         \multirow{4}{*}{\makecell{ResNet50/\\ImageNet/\\fine-tuned}} & Accuracy & 85.10\% & 85.26\% \\
         & Precision & 76.09\% & 79.89\% \\
         & Recall & \textbf{76.21\%} & 75.17\% \\
         & F1-Score & 75.38\% & 75.79\% \\
         \hline
         \multirow{4}{*}{\makecell{ResNet50/\\ImageNet/\\frozen}} & Accuracy & 82.90\% & 86.37\% \\
         & Precision & 74.22\% & 80.75\% \\
         & Recall & 73.30\% & 76.00\% \\
         & F1-Score & 71.30\% & 77.32\% \\
         \hline
         \multirow{4}{*}{\makecell{DINOv2/\\SoftCon/\\fine-tuned}} & Accuracy & \textbf{78.75\%} & 61.69\% \\
         & Precision & \textbf{64.28\%} & 24.01\% \\
         & Recall & \textbf{67.69\%} & 44.44\% \\
         & F1-Score & \textbf{62.97\%} & 31.16\% \\
         \hline
         \multirow{4}{*}{\makecell{DINOv2/\\SoftCon/\\frozen}} & Accuracy & 79.51\% & 84.32\% \\
         & Precision & 66.40\% & 76.69\% \\
         & Recall & 70.03\% & 73.94\% \\
         & F1-Score & 63.26\% & 71.96\% \\
         \hline
         \multirow{4}{*}{\makecell{DINOv2/\\ImageNet/\\fine-tuned}} & Accuracy & \textbf{77.98\%} & 61.95\% \\
         & Precision & \textbf{62.99\%} & 30.75\% \\
         & Recall & \textbf{68.30\%} & 55.56\% \\
         & F1-Score & \textbf{62.95\%} & 39.56\% \\
         \hline
         \multirow{4}{*}{\makecell{DINOv2/\\ImageNet/\\frozen}} & Accuracy & \textbf{79.96\%} & 75.04\% \\
         & Precision & \textbf{71.20\%} & 63.29\% \\
         & Recall & \textbf{67.85\%} & 63.44\% \\
         & F1-Score & \textbf{63.17\%} & 56.99\% \\    
         \hline
    \end{tabular}
    \end{adjustbox}
    \caption{Model performance comparison}
    \label{tab:results}
\end{table}


\section{Discussion}
As shown in Table~\ref{tab:results}, the impact of PCA on classification performance varies significantly between ResNet50 and DINOv2 models, as well as between fine-tuned and frozen encoder configurations.

For ResNet50-based models, PCA has a predominantly negative effect. In all configurations, models trained without PCA achieve higher accuracy and F1-scores, suggesting that PCA removes useful spectral-spatial information which convolutional layers naturally extract. The performance decline is especially pronounced in frozen encoder models, where the inability to adapt to the lower-dimensional input leads to poorer results. Likewise, PCA reduces accuracy and precision in fine-tuned configurations, indicating that it disrupts the feature extraction process essential for CNN-based models.

By contrast, DINOv2-based models exhibit the opposite trend. Fine-tuned models benefit considerably from PCA, attaining notable improvements in accuracy and F1-score. This suggests that PCA improves feature separability by minimising redundant spectral information. However, when the encoder remains frozen, PCA negatively affects performance, as the fixed feature representations cannot adapt to the transformed input, resulting in diminished classification effectiveness.

\section{Conclusion}

This paper examined the impact of Principal Component Analysis (PCA) on multi-label classification of multispectral imagery using ResNet50 and DINOv2 encoders. Our findings reveal that PCA affects different architectures in distinct ways.

For CNN-based models (ResNet50), PCA led to a consistent decline in classification performance, particularly in configurations where encoder weights were frozen. This suggests that convolutional networks already capture essential spectral-spatial features, and reducing input dimensionality removes critical information required for accurate classification.

By contrast, when fine-tuned, transformer-based models (DINOv2) showed significant improvements with PCA, yielding substantial gains in accuracy and F1-score. PCA likely helps eliminate redundant spectral information, allowing DINOv2 to better separate features within the learned representation space. However, in frozen encoder configurations, PCA degraded performance, likely due to the model’s inability to adapt to the transformed input. The encoder weights were pre-trained on the original thirteen-channel dataset and thus may be unable to utilise the information compressed with PCA properly.

Additionally, as can be seen in Tables~\ref{tab:inference} and \ref{tab:size_comprehension}, PCA yielded a smaller model size and efficiency gains in inference time for ResNet50. This indicates that another advantage of PCA lies in computational efficiency, especially when processing large datasets. It also greatly improves training times for these large models.

Overall, our results indicate that PCA’s effectiveness is highly architecture-dependent. While it provides little benefit for ResNet50 in terms of accuracy, it can enhance fine-tuned transformer models such as DINOv2. Nevertheless, applying PCA to frozen encoders appears counterproductive, disrupting pretrained feature distributions.

Perhaps the benefits of using PCA for dimensionality reduction would be more prominent if the SoftCon training of the encoder were performed on the dataset already processed by PCA, rather than on the high-dimensional raw data. That would allow the feature extractor to take advantage of high-variance, low-noise data and potentially improve feature extraction further. Future research could also explore alternative dimensionality reduction techniques, such as autoencoders, t-SNE, or UMAP, to determine whether they provide better feature preservation and classification performance than PCA. Additionally, experimenting with hybrid models combining CNNs and transformers could offer new insights into leveraging local and global feature extraction for improved accuracy. Furthermore, evaluating the impact of dimensionality reduction techniques on different data types and domains beyond multispectral imagery could provide valuable insights into their broader applicability. For instance, testing PCA and alternative methods in medical imaging, financial time-series analysis, or natural language processing tasks could reveal whether similar trends in feature extraction efficiency hold across various data structures and modalities. Extending such studies could contribute to a more generalised understanding of how dimensionality reduction affects deep learning performance across diverse fields.


\section*{Acknowledgments}
The research was funded by POB Cybersecurity and Data Analysis of Warsaw University of Technology within the Excellence Initiative: Research University (IDUB) program.

\printbibliography

\end{document}